\documentclass[letterpaper]{article} 
\usepackage{aaai25}  
\usepackage{times}  
\usepackage{helvet}  
\usepackage{courier}  
\usepackage[hyphens]{url}  
\usepackage{graphicx} 
\usepackage{natbib}  
\usepackage{caption} 
\usepackage{algorithm}
\usepackage{algorithmic}
\usepackage{amsmath}  
\usepackage{amsfonts}     
\usepackage{amssymb}      
\usepackage{booktabs} 
\usepackage{multirow}
\usepackage{array}

\usepackage{newfloat}
\usepackage{listings}
\DeclareCaptionStyle{ruled}{labelfont=normalfont,labelsep=colon,strut=off} 
\floatstyle{ruled}
\newfloat{listing}{tb}{lst}{}
\floatname{listing}{Listing}
\title{RSEA-MVGNN: Multi-View Graph Neural Network with Reliable Structural Enhancement and Aggregation}
\author{
    Junyu Chen\textsuperscript{\rm 1},
    Long Shi\textsuperscript{\rm 1},
    Badong Chen\textsuperscript{\rm 2}
}
\affiliations{
    \textsuperscript{\rm 1}Financial Intelligence and Financial Engineering Key Laboratory of Sichuan Province, \\
    School of Computing and Artificial Intelligence, Southwestern University of Finance and Economics\\
    \textsuperscript{\rm 2}Institute of Artificial Intelligence and Robotics, Xi’an Jiaotong University\\
    223081200039@smail.swufe.edu.cn, shilong@swufe.edu.cn, chenbd@mail.xjtu.edu.cn
}

\usepackage{bibentry}

\begin{document}

\maketitle

\begin{abstract}
Graph Neural Networks (GNNs) have exhibited remarkable efficacy in learning from multi-view graph data. In the framework of multi-view graph neural networks, a critical challenge lies in effectively combining diverse views, where each view has distinct graph structure features (GSFs). 
Existing approaches to this challenge primarily focus on two aspects: 1) prioritizing the most important GSFs, 2) utilizing GNNs for feature aggregation. However, prioritizing the most important GSFs can lead to limited feature diversity, and existing GNN-based aggregation strategies equally treat each view without considering view quality. 
To address these issues, we propose a novel Multi-View Graph Neural Network with Reliable Structural Enhancement and Aggregation (RSEA-MVGNN). 
Firstly, we estimate view-specific uncertainty employing subjective logic. Based on this uncertainty, we design reliable structural enhancement by feature de-correlation algorithm. This approach enables each enhancement to focus on different GSFs, thereby achieving diverse feature representation in the enhanced structure. 
Secondly, the model learns view-specific beliefs and uncertainty as opinions, which are utilized to evaluate view quality. Based on these opinions, the model enables high-quality views to dominate GNN aggregation, thereby facilitating representation learning.
Experimental results conducted on five real-world datasets demonstrate that RSEA-MVGNN outperforms several state-of-the-art GNN-based methods.
\end{abstract}

%

\section{Introduction}
Graphs are extensively employed for constructing and interpreting data that encompasses intricate interconnections. For example, in urban transportation networks \cite{transportation-}, data is naturally depicted as a graph, with nodes symbolizing urban transportation stations and edges denoting the flow of traffic between nodes. 
In the analysis of traffic flow, each urban transportation station can collect data from different temporal scales, such as weekdays, weekends and monthly periods. This indicates that each node could be associated with multiple data sources. These characteristics spawn the emergence of an advanced graph modelling method, namely multi-view graph, which contains consistent nodes but shows different structures in each view.
Multi-view graph representation learning \cite{repre1-, repre2-, repre3-} aims to merge information from various views into a compact, high quality representation. This area is garnering increasing research interest, and has found extensive applications in traffic analysis \cite{Citywide-}, disease diagnosis \cite{Pathology-}, protein prediction \cite{proteins-}, and among others.

\begin{figure}[t]
\centering
\includegraphics[width=1\columnwidth]{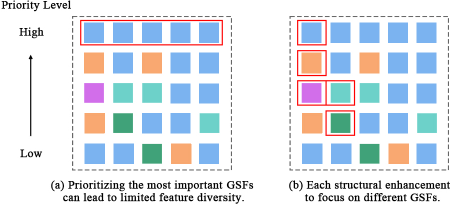} 
\caption{Visualization of the GSFs selected during structure enhancement. Squares of different colors represent nodes with different features in graphs. The red outlines indicate the enhanced nodes.}
\label{fig1}
\end{figure}

Early multi-view graph neural networks (MVGNNs) directly utilize GNNs for multi-view aggregation, such as Graph Convolutional Networks (GCNs) \cite{G2-} and Graph Attention Networks (GATs) \cite{HAN-}, focusing on the architectural design of MVGNNs to facilitate aggregation. Subsequently, MVGNNs evolve into more advanced frameworks \cite{L2-, repre2-, T2-}, which first prioritize view-specific graph structure features (GSFs) and then employ GNNs for aggregation. This approach enables MVGNNs to focus on important GSFs, thereby achieving more effective fusion. 

In terms of GSFs processing, \cite{L2-} augments important features to construct latent graphs, \cite{T2-} selects the most crucial neighboring nodes to form an enhanced weight matrix, and \cite{mvgnn_3-} purifies key features to enrich separate views. Under this mechanism of prioritizing the most important GSFs, it inevitably leads to the result shown in Fig. \ref{fig1} (a), namely limited feature diversity in the enhanced structure. Due to the fact of neglecting diverse GSFs information, this approach is unable to capture a more comprehensive representation.
In terms of aggregation, various types of GNN-based methods \cite{mvgnn_1-, MAGNN-, MGNN1-} assume that different views have equal quality, thus all views participate equally in the aggregation process. This assumption is unreasonable, as the qualities of multiple views are different in practical scenes \cite{reliable-, shi2024enhanced}. Treating all views equally limits the positive contributions of high-quality views while allowing the negative impacts of low quality views on representation learning.

\begin{figure*}[t]
\centering
\includegraphics[width=1\textwidth]{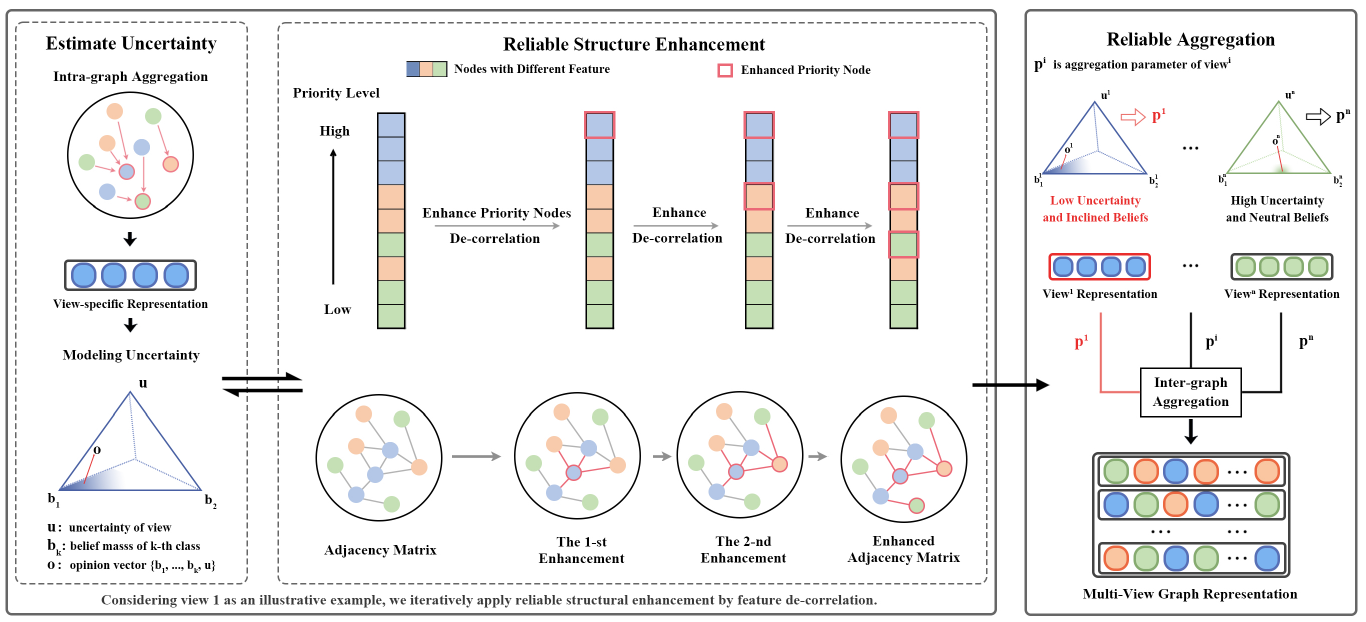} 
\caption{Illustration of RSEA-MVGNN. First, we learn view-specific beliefs and uncertainty as opinions. Based on the uncertainty, we apply reliable structural enhancement by feature de-correlation. Second, we construct aggregation parameters based on opinions of enhanced views, utilizing these parameters to facilitate high-quality views dominating inter-graph aggregation.}
\label{fig2}
\end{figure*}

To address the above discussed issues, we propose the Multi-View Graph Neural Network with Reliable Structural Enhancement and Aggregation (RSEA-MVGNN). As illustration in Fig. \ref{fig2}, this framework includes two components: reliable structural enhancement and reliable aggregation.
1) Based on subjective logic theory \cite{trusted1-, trusted2-}, we estimate view-specific uncertainty. The uncertainty reflects the degree of support provided by the view-specific predictions. To achieve reliable structural enhancement, we continue the enhancement process when uncertainty decreases after each iteration, and terminate it otherwise. The process employs a feature de-correlation algorithm for each enhancement. This approach enables focusing on different GSFs, thereby achieving diverse feature representation in the enhanced structure.
2) For reliable aggregation in MVGNN, we learn view-specific opinions consisting of belief masses and uncertainty. To evaluate view quality, we construct aggregation parameters based on these opinions. High-quality views have inclined category beliefs and lower uncertainty, resulting in larger aggregation parameters. Our model utilizes the parameters to enable high-quality views to dominate GNN aggregation, thereby achieving better multi-view graph representation learning.

Key contributions: 
1) Reliable structural enhancement, utilizing uncertainty as a criterion to ensure more reliable enhancement results, and employing feature de-correlation to enrich the diversity of features in enhanced structures. 
2) In the reliable aggregation, we utilize aggregation parameters to guide the fusion among multiple views. This approach enhances the positive contributions of high-quality views while limiting the negative impacts of low-quality views, thereby improving the performance of the multi-view graph aggregation.
3) The RSEA-MVGNN outperforms several state-of-the-art baselines. When compared to the existing top-performing method, it achieves a remarkable increase of 13.91\% in the Ma-F1 score for classification tasks. Moreover, for clustering tasks, it obtains an improvement of 15.05\% in the Adjusted Rand Index (ARI).

\section{Related Work}
\subsubsection {Multi-View Graph Neural Networks:} MVGNNs primarily focus on two crucial aspects: extracting GSFs and aggregating information from multi-view graphs. The extraction of GSFs determines the quality used in subsequent aggregation. Extracting GSFs has two main categories: Rule-based methods and adaptive methods. Rule-based methods \cite{S1-, S3-, GSN-} rely on predefined rules or domain knowledge to extract GSFs. These methods extract features based on graph topology, such as node degree \cite{node-} and centrality measures \cite{mvgnn_2-}. Adaptive methods \cite{R1-, R2-} provide flexibility for enhancing critical GSFs. A typical example \cite{T2-} utilizes reinforcement learning to select the relevant neighborhoods. However, emphasizing the important GSFs may result in overlooking the diversity of GSFs and fail to capture a more comprehensive representation. Therefore, we design a reliable structural enhancement by feature de-correlation for ensuring diverse GSFs.

In terms of aggregation, MVGNNs primarily focus on neural network design. Specifically, GCN-based methods \cite{G2-} unify GCNs and co-training into a single framework, enabling interaction fusion between views. GAT-based methods \cite{HAN-} employ hierarchical attention mechanisms that include node-level and semantic-level attentions. These methods treat all views equally, disregarding their different quality. However, each view has different quality in real-world scenarios. To tackle this limitation, we propose a reliable aggregation method based on view-specific opinions.


\subsubsection {Uncertainty-based Deep Learning:} 
In practice, a crucial challenge lies in handling varying quality of multimodal data \cite{lowsur-}. To address this issue, uncertainty-based deep learning has emerged as a more general and principled approach for reliable multimodal fusion. 
Specifically, Evidential Deep Learning (EDL) \cite{trusted1-} calculates category-specific evidence using a single deep neural network. Building on EDL, Trusted Multi-View Classification \cite{SL2-} combines subjective logic and Dempster-Shafer theory to perform reliable fusion of multi-view. Furthermore, \cite{reliable-} introduced a conflictive opinion aggregation strategy and provided theoretical proof that uncertainty increases for conflictive instances. Similarly, quantifying uncertainty in GNNs has gained significant attention. \cite{ungraph-} categorizes existing uncertainty quantification methods into single deterministic model \cite{UG_1-} and single model with random parameters \cite{UG_2-}. However, there is limited research exploring reliable aggregation in MVGNNs. Building on this observation, our research bridges the gap between multi-view uncertainty learning and GNN aggregation. 

\section{Our Method}
This section begins with an introduction to MVGNNs and subjective logic preliminaries. Subsequently, we discuss the process of reliable structural enhancement and aggregation. 

\subsection{Preliminaries}

\subsubsection{Multi-view Graph Representation Learning.}
Let $\mathcal{G}$ denote the collection of $M$ distinct multi-view graphs, expressed as $\mathcal{G} = \{\mathcal{G}_i\}^M_{i=1}$. For any multi-view graph $\mathcal{G}_i$, it can be represented by $\mathcal{G}_i = \{\mathbf{G}_{i,j}\}^V_{j=1}$, where $\mathbf{G}_{i,j}$ denotes the graph corresponding to the $j$-th view within $\mathcal{G}_i$. In each graph $\mathbf{G}_{i,j}=\{N,E,\mathbf{A},\mathbf{F}\}$, the set of nodes is $N = \{n_k\}^{|N|}_{k=1}$, and the set of edges is denoted by $E\subseteq N\times N$. The weighted adjacency matrix $\mathbf{A}\in \mathbb{R}^{|N|\times |N|}$, with $\mathbf{A}(k,k')$ quantifying the connection weight between node $n_k$ and node $n_{k'}$. The feature matrix $\mathbf{F}\in \mathbb{R}^{|N|\times D}$, captures the node features across $|N|$ nodes and $D$ dimensions, where $\mathbf{F}(k)$ represents the feature vector for node $n_k$.
For multi-view graph representation learning, the objective is to embed the multi-view graphs $\mathcal{G} = \{\mathcal{G}_i\}^M_{i=1}$ into a low-dimensional and high quality feature matrix $\mathbf{Z}\in \mathbb{R}^{|M|\times D^{(l)}}$, where $D^{(l)}$ represents the dimension of feature at the $l$-th layer of representation learning process. Each $\mathbf{Z}_i$ denotes the feature representation vector corresponding to a multi-view graph $\mathcal{G}_i$.

\subsubsection{Multi-view Graph Neural Network.}
This GNN framework is tailored for learning representations of multi-view graphs. Given a multi-view graph $\mathcal{G}_i$, the $j$-th view is represented by $\mathbf{G}_{i,j}$, which includes an adjacency matrix $\mathbf{A}_{i,j}$ and a feature matrix $\mathbf{F}_{i,j}$. At the $l$-th layer for $j$-th view, the intra-graph feature fusion process is defined by the following equation:
\begin{equation}
\begin{aligned}
\mathbf{F}^{(l^*)}_{i,j} &= \sigma \left( GNN^{(l)}_{intra} \left( \mathbf{F}^{(l-1)}_{i,j}, \mathbf{A}^{(l-1)}_{i,j}  \right) \right),
\end{aligned}
\label{intra-gnn}
\end{equation}
where $\mathbf{F}^{(l^*)}_{i,j}$ represents the intermediate feature matrix between the $(l-1)$-th and $l$-th layer, $GNN^{(l)}_{intra}(\cdot)$ is the intra-graph fusion function at the $l$-th layer, and $\sigma(\cdot)$ denotes the activation function.

Subsequently, the inter-graph feature fusion process for the multi-view GNN at the $l$-th layer can be expressed as:
\begin{equation}
\begin{aligned}
\mathbf{F}^{(l)}_{i,j} &= \sigma \left( GNN^{(l)}_{inter} \left( \left\{ \mathbf{F}^{(l^*)}_{i,j} \right\}^{V}_{j=1} \right) \right),
\end{aligned}
\label{inter-gnn}
\end{equation}
where $GNN^{(l)}_{inter}(\cdot)$ is the function for inter-graph fusion at the $l$-th layer, integrating graph features from different views.

\subsubsection{Subjective Logic.}
Subjective logic provides a theoretical framework \cite{SL1-, trusted2-} that associates the parameters of the Dirichlet distribution with the evidence from each view. This framework quantifies the overall uncertainty of the results from each view and reflects the reliability of the view.

The Dirichlet distribution is dependent on the parameters \( \boldsymbol{\alpha} = [\alpha_1, \ldots, \alpha_K] \), which dictate the shape of the distribution. The Dirichlet Probability Density Function is defined as:
\begin{equation}
\begin{aligned}
D(\mathbf{p} | \boldsymbol{\alpha}) = 
\begin{cases} 
\frac{1}{B(\boldsymbol{\alpha})} \prod_{k=1}^{K} p_k^{\alpha_k - 1}, & \text{for } \mathbf{p} \in S_K, \\
0, & \text{otherwise},
\end{cases} 
\end{aligned}
\label{dirichlet}
\end{equation}
where the set \( S_K \) is the $K$-dimensional unit simplex, which defines the domain of the probability vector \( \mathbf{p} = [p_1, \ldots, p_K]^T \). Additionally, \( B(\boldsymbol{\alpha}) \) is the $K$-dimensional multinomial beta function.

In the subjective logic framework, for a given view $\mathbf{G}_{i,j}$, the evidence vector $\mathbf{e}_{i,j} = [e_1^{i,j}, \ldots, e_K^{i,j}]$ is the classification result of a view-specific neural network. This vector $\mathbf{e}_{i,j}$ is applied to derive the belief vector $\mathbf{b}^{i,j} = [b_1^{i,j}, \ldots, b_K^{i,j}]$, which reflects the reliability assigned to each of the $K$ classes. Simultaneously, the uncertainty mass $u^{i,j}$ is the total uncertainty in the evidence vector. According to subjective logic, both $\mathbf{b}^{i,j}$ and $u^{i,j}$ are required to be non-negative, and their sum must equal one:
\begin{equation}
\sum_{k=1}^{K} b_k^{i,j} + u^{i,j} = 1, \quad \forall k \in [1, \ldots, K],
\label{bu1}
\end{equation}
where $b_k^{i,j} \geq 0$ and $u^{i,j} \geq 0$. 

In the $j$-th view, subjective logic associates the evidence $\mathbf{e}_{i,j}$, to the Dirichlet distribution parameters, \( \boldsymbol{\alpha}_{i,j} = [\alpha_1^{i,j}, \ldots, \alpha_K^{i,j}] \). Specifically, each $\alpha_k^{i,j}$ is calculated by $\alpha_k^{i,j} = e_k^{i,j} + 1$. Following this approach, the belief mass $b_k^{i,j}$ and the uncertainty $u^{i,j}$ are determined by the formulas:
\begin{equation}
b_k^{i,j} = \frac{e_k^{i,j}}{S^{i,j}} = \frac{\alpha_k^{i,j} - 1}{S^{i,j}} \quad \text{and} \quad u^{i,j} = \frac{K}{S^{i,j}},
\label{bu2}
\end{equation}
where $S^{i,j} = \sum_{k=1}^{K} (e_k^{i,j} + 1) = \sum_{k=1}^{K} \alpha_k^{i,j}$ is the Dirichlet strength. Consequently, by utilizing the Dirichlet distribution, subjective logic enables the modeling of both second-order probabilities and uncertainty in neural network outputs \cite{SL2-}.

\subsection{Reliable Structural Enhancement and Aggregation}
\subsubsection{Estimate Uncertainty of Intra-GNN.}
To estimate the uncertainty in intra-graph feature fusion, we first employ a Graph Convolutional Network. This process is described by the following equation:
\begin{equation}
\begin{aligned}
\mathbf{F}^{(l^*)}_{i,j} &= \sigma \left( \mathbf{A}^{(l-1)}_{i,j} \mathbf{F}^{(l-1)}_{i,j} \mathbf{W}^{(l)}_{intra,i,j} \right),
\end{aligned}
\label{midEvid}
\end{equation}
where $\mathbf{W}^{(l)}_{intra,i,j}$ is the weight matrix corresponding to the intra-graph fusion at $l$-th layer. Building on this, we utilize a single-layer feedforward neural network (FNN) to process the result of the intra-graph fusion. The FNN generates the evidence $\mathbf{e}_{i,j}$ for $j$-th view, specifically:
\begin{equation}
\begin{aligned}
\mathbf{e}_{i,j} &= \hat{\sigma} \left( \mathbf{F}^{(l^*)}_{i,j} \mathbf{W}^{(l)}_{fnn,i,j} \right),
\end{aligned}
\label{evid}
\end{equation}
where $\mathbf{W}^{(l)}_{fnn,i,j}$ is the weight matrix of the FNN, $\hat{\sigma}$ refers to the softplus activation function. The softplus function is employed to ensure non-negative network output, which is necessary for acquiring the parameters of the Dirichlet distribution. In the subjective logic framework, following Eq. (\ref{bu2}) from the preliminaries, the total uncertainty $u^{i,j}$ is calculated based on the evidence vector $\mathbf{e}_{i,j}$.

\subsubsection{Reliable Structural Enhancement.}
Fig. \ref{fig2} shows the process for reliable structural enhancement. To enhance structures with diverse features and reduce uncertainty, we apply reliable structural enhancement by feature de-correlation. This algorithm enables each structural enhancement to focus on different GSFs.
Algorithm \ref{alg1} is designed to enhance structure while performing feature de-correlation. Specifically, for a graph $\mathbf{G}_{i,j}$, we apply degree centrality to its adjacency matrix $\mathbf{A}_{i,j}$ to measure the influence of nodes in the graph. Concurrently, we analyze the variance of node features in the matrix $\mathbf{F}_{i,j}$, quantifying the diverse feature information exhibited by the nodes.
For $k$-th node, the node priority level is calculated by:
\begin{equation}
    \phi_k = CEN(\mathbf{A}_{i,j}(k)) + VAR(\mathbf{F}_{i,j}(k)),
\end{equation}
where $\mathbf{A}_{i,j}(k)$ and $\mathbf{F}_{i,j}(k)$, represent the adjacency vector and feature vector of $k$-th node. The degree centrality and variance are calculated using $CEN(\cdot)$ and $VAR(\cdot)$, respectively. 
The priority level vector $\mathbf{\Phi} = [\phi_1, \ldots, \phi_{|N|}]$ represents the importance of nodes in the view $\mathbf{G}_{i,j}$, where higher values indicate higher priority.

\begin{algorithm}[tb]
\caption{Reliable Structural Enhancement by Feature De-correlation}
\label{alg1}

\textbf{Input}: The adjacency matrix $\mathbf{A}_{i,j}$, the feature matrix $\mathbf{F}_{i,j}$\\
\textbf{Output}: The enhanced adjacency matrix $\hat{\mathbf{A}}_{i,j}$

\begin{algorithmic}[1]
\STATE Initialize uncertainty $U = \infty$
\STATE Initialize enhancement number $R = 0$
\STATE Iteration factor $T = 0.05|N|$

\STATE Calculate degree centrality vector $\mathbf{\Delta} = [\delta_1, \ldots, \delta_{|N|}]$ based on the adjacency matrix $\mathbf{A}_{i,j}$, where $\delta_i$ is the degree centrality of the $i$-th node
\STATE Calculate variance vector $\mathbf{\Theta} = [\theta_1, \ldots, \theta_{|N|}]$ based on the feature matrix $\mathbf{F}_{i,j}$, where $\theta_i$ is the variance of features for the $i$-th node
\STATE Calculate the priority level vector $\mathbf{\Phi} = \mathbf{\Delta} + \mathbf{\Theta}$

\WHILE{$True$}
    \STATE Intermediate feature $\mathcal{F}^{(l^*)}(:,j,:,:)$ by Eq. (\ref{midEvid})
    \STATE Uncertainty $U'$ by Eq. (\ref{bu2}) and (\ref{evid})
    
    \IF {$U'<=U$}
    \STATE Update uncertainty $U = U'$
    \STATE Update enhancement number $R = R + T$
    \FOR{$iter = 1$ to $R$}
        \STATE Find the priority index $Ind = \arg\max_k \phi_k$
        \STATE Enhance the priority $Ind$-th node by Eq. (\ref{find})
        \STATE Update $\mathbf{\Phi}$ by applying the feature de-correlation process according to Eq. (\ref{decor})
    \ENDFOR
    \ELSE
    \STATE $Break$
    \ENDIF
\ENDWHILE

\end{algorithmic}
\end{algorithm}

The node with the highest value in $\mathbf{\Phi}$ is selected as the priority node, denoted as the $Ind$-th node. We obtain the priority adjacency vector $\mathbf{A}_{i,j}(Ind)$, where each positive weight represents an effective connection to other nodes. To construct the adjacency vector mask for the $Ind$-th node, denoted by $\mathbf{\widetilde{A}}_{i,j}(Ind)$, we assign one to each positive weight in $\mathbf{A}_{i,j}(Ind)$, and zero to all others. The priority adjacency vector $\mathbf{A}_{i,j}(Ind)$ is enhanced by the following equation:
\begin{equation}  
    \mathbf{\hat{A}}_{i,j}(Ind) = max(\mathbf{A}_{i,j})  \mathbf{\widetilde{A}}_{i,j}(Ind),
\label{find}
\end{equation}
where $max(\mathbf{A}_{i,j})$ serves to select the maximum edge weight in the adjacency matrix $\mathbf{A}_{i,j}$. The enhanced adjacency vector $\mathbf{\hat{A}}_{i,j}(Ind)$ means the edges with positive weights of the $Ind$-th node are enhanced to $max(\mathbf{A}_{i,j})$. By enhancing the edge weights of the $Ind$-th node, we increase its connectivity within the network, thereby improving its influence in the propagation process of the GNN.

To ensure diversity in subsequently enhanced priority nodes, we reduce the priority level of nodes with similar features. The priority level of $\mathbf{\phi}_k$ is updated by:
\begin{equation}
    \mathbf{\phi}_k = \mathbf{\phi}_k (1-COS(\mathbf{F}_{i,j}(k),\mathbf{F}_{i,j}(Ind)))
\label{decor}
\end{equation}
where $COS(\cdot)$ measures the cosine similarity between the node feature vectors $\mathbf{F}_{i,j}(k)$ and $\mathbf{F}_{i,j}(Ind)$. The priority level vector $\mathbf{\Phi}$ is updated for each node by Eq. (\ref{decor}). 

Following Algorithm \ref{alg1}, we alternately estimate uncertainty and enhance structure through feature de-correlation. If uncertainty decreases after each enhancement, we increase the number of enhanced nodes for the next enhancement. We continue the enhancement process when uncertainty decreases after each iteration, and terminate it otherwise. After reliable structural enhancement, we obtain the enhanced adjacency matrix $\hat{\mathbf{A}}_{i,j}$, which exhibits lower uncertainty and greater diversity in features representation.

\subsubsection{Reliable Aggregation.} 
To achieve reliable aggregation, we measure view-specific opinions consisting of belief vector $\mathbf{b}^{i,j}$ and uncertainty mass $u^{i,j}$. Based on opinions, we construct the aggregation parameter to evaluate view quality. 
The aggregation parameter $p_{i,j}$ is calculated by:
\begin{equation}
    p_{i,j} =  VAR(\mathbf{b}^{i,j})/u^{i,j}
\end{equation}
where variance is calculated using $VAR(\cdot)$. High-quality views have inclined category beliefs and lower uncertainty, resulting in larger aggregation parameters. For example, under a triple classification task, given $\mathbf{b}^{i,0} = [0.4, 0.1, 0.1]$ and $\mathbf{b}^{i,1} = [0.2, 0.2, 0.2]$, the $\mathbf{b}^{i,0}$ have larger variance and more inclined category beliefs, which reveal more meaningful insights. The uncertainty $u^{i,j}$ is inverted to represent reliability and serves as the denominator. Our model utilizes the aggregation parameters to enable high-quality views to dominate the inter-graph aggregation. 

In the inter-graph feature aggregation process, we guide reliable aggregation by utilizing the aggregation parameter vector $\mathbf{p}_i = [p_{i,1}, \ldots, p_{i,V}]$. To align with the dimensions of intermediate feature tensor $TRAN{1,2}(\mathcal{F}^{(l^*)}_i)$, we span $\mathbf{p}_i$ and denote the result as $\mathcal{P}_i$. Here, $\mathcal{P}_i$ represents the quality of multiple views. We define the inter-graph reliable aggregation process:
\begin{equation}
    TRAN_{1,2} \left( \mathcal{F}^{(l)}_i \right) = \sigma \left(\mathcal{P}_i \odot TRAN_{1,2} \left( \mathcal{F}^{(l^*)}_i \right) \mathcal{W}^{(l)}_{inter,i} \right),
\label{inter}
\end{equation}
where $\mathcal{W}^{(l)}_{inter,i}$ is the weight tensor for the inter-graph aggregation of $i$-th view in $l$-th layer, and operation $\odot$ denotes the Hadamard product. Based on $\mathcal{P}_i$, each view is participates in the inter-graph aggregation according to its quality. Consequently, views with higher quality have a greater influence on the inter-graph aggregation process. Our research bridges the gap between multi-view uncertainty learning and GNN aggregation. The reliable aggregation method differentially treats views of varying quality. This novel approach enhances the positive contribution of reliable views while mitigating the impact of less reliable ones, achieving reliable multi-view graph aggregation.

In the final layer, we vectorize $\mathcal{F}^{(l)}_i$ to obtain the result of our multi-view graph representation learning, represented by $\mathbf{Z}_i \in \mathbb{R}^{1 \times D^{(l)}}$. This process is implemented using mean pooling:
\begin{equation}
    \mathbf{Z}_i = \frac{1}{V|N|} \sum_{j=1}^{V} \sum_{k=1}^{|N|} \mathcal{F}^{(l)}(i,j,k,:),
\end{equation}
where $\mathbf{Z}_i$ denotes the $i$-th row of the feature representation matrix $\mathbf{Z}$, corresponding to the $i$-th multi-view graph instance $\mathcal{G}_i$.

\subsection{Optimization}
In the design of the loss function, based on the theory of subjective logic, we obtain the parameter $\boldsymbol{\alpha}$ and form the Dirichlet Probability Density Function $D(\mathbf{p} | \boldsymbol{\alpha})$, where $\mathbf{p}$ is the class assignment probabilities on a simplex. We formulate the adjusted cross-entropy loss function \cite{trusted2-}: 
\begin{equation}
\begin{aligned} 
    \mathcal{L}_{ace}(\boldsymbol{\alpha}) &= \int \left[ \sum_{k=1}^{K} -Y_{k} \log{(p_k)} \right] \frac{\prod_{k=1}^{K} p_k^{\alpha_{k} - 1}}{B(\boldsymbol{\alpha})} d \mathbf{p} \\
    &= \sum_{k=1}^{K} Y_k (\psi(S) - \psi(\alpha_k)),
\end{aligned}
\label{ls1}
\end{equation}
where $Y_k$ is the actual label and $p_k$ is the predicted probability for the k-th class. The strength $S$ of Dirichlet distribution is given by Eq. (\ref{bu2}), and the term $\psi(\cdot)$ is defined as the digamma function. Eq. (\ref{ls1}) represents the integral of the cross-entropy loss function over the simplex defined by $\boldsymbol{\alpha}$. 

The overall loss of RSEA-MVGNN is defined as follows:
\begin{equation}
    \mathcal{L}_{all} = \sum_{l=1}^{L} \left[ \mathcal{L}_{ace}(\boldsymbol{\alpha}^{(l)}) + \sum_{j=1}^{V} \mathcal{L}_{ace}(\boldsymbol{\alpha}_j^{(l)})  + \lambda \left\| \Theta^{(l)} \right\|_2 \right],
\end{equation}
where $\boldsymbol{\alpha}_j^{(l)}$ and $\boldsymbol{\alpha}^{(l)}$ are respectively constructed from the evidence formulated by Eqs. (\ref{midEvid}) and (\ref{inter}). To mitigate the risk of overfitting, we introduce a regularization term $\Theta^{(l)}$, with $\lambda$ serving as the regularization coefficient. 

\section{Experiments}
In this section, we conduct comprehensive experiments to evaluate the performance of the proposed RSEA-MVGNN on five domain-specific datasets. Our evaluation includes a diverse set of tasks including classification, clustering, ablation studies, computational complexity analysis. More details are shown in the appendix.
\begin{table}[htbp]
\centering
\small 
\setlength{\tabcolsep}{0.2mm} 
\begin{tabular}{>{\centering\arraybackslash}m{1.7cm}>{\centering\arraybackslash}m{1.2cm}>{\centering\arraybackslash}m{1cm}>{\centering\arraybackslash}m{0.65cm}>{\centering\arraybackslash}m{4cm}}
\toprule
{\small Dataset} & {\small Instances} & {\small Classes} & {\small Features} & {\small Views} \\
\midrule
HIV & 70 & 2 & 90 & fMRI\&DTI \\
BikeDC & 72 & 4 & 267 & {\small weekday\&weekend\&month} \\
{\small PROTEINS} & 1000 & 2 & 80 & {\small sequence\&molecule interaction} \\
{\small ogbg-molhiv} & 41127 & 2 & 9 & self-duplicate \\
ACM & 3025 & 3 & 1830 & APA\&APSPA \\
\bottomrule
\end{tabular}
\caption{Datasets Statistics}
\label{datasets}
\end{table}

\begin{table*}[!ht]
    \centering
    \small 
    \caption{Experimental results (\%) for classification and clustering tasks on datasets.}
    \label{expdata1}
    \begin{tabular}{cccccccc}
    \toprule
        \multirow{2}{*}{Datasets} & Metrics & \multicolumn{2}{c}{Ma-F1} & \multicolumn{2}{c}{Mi-F1} & \multirow{2}{*}{NMI} & \multirow{2}{*}{ARI} \\
        ~ & Train\% & 20\% & 60\% & 20\% & 60\% & ~ & ~ \\ \midrule
        \multirow{8}{*}{HIV} & HAN & 58.90 ± 5.04 & 69.55 ± 4.23 & 60.00 ± 3.64 & 70.62 ± 4.88 & 23.73 ± 7.17 & 17.59 ± 14.75 \\ 
        ~ & MAGNN & 56.66 ± 4.75 & 69.84 ± 4.59 & 58.43 ± 3.14 & 71.25 ± 5.00 & 19.30 ± 10.96 & 20.33 ± 11.98 \\ 
        ~ & TensorGCN & 59.02 ± 4.38 & 70.98 ± 4.36 & 60.31 ± 2.81 & 72.50 ± 5.00 & 24.38 ± 13.50 & 19.44 ± 9.14 \\ 
        ~ & RTGNN  & 65.74 ± 5.75 & 75.14 ± 8.18 & 67.50 ± 6.43 & 76.25 ± 08.29 & 35.02 ± 11.43 & 30.86 ± 10.81 \\ 
        ~ & PTGB & 68.33 ± 6.78 & 81.74 ± 8.12 & 70.24 ± 8.04 & 82.57 ± 8.43 & 36.24 ± 9.35 & 34.78 ± 9.49 \\
        ~ & AdaSNN & 62.56 ± 6.37 & 72.48 ± 6.96 & 63.88 ± 6.47 & 73.22 ± 6.84 & 34.57 ± 8.61 & 34.67 ± 9.01 \\        
        ~ & RSEF-MVGNN & \textbf{70.58 ± 3.82} & \textbf{83.43 ± 5.38} & \textbf{72.34 ± 3.01} & \textbf{84.38 ± 5.03} & \textbf{49.31 ± 15.92} & \textbf{49.83 ± 22.40} \\ 
        ~ & Gain & 2.25 & 1.69 & 2.10 & 1.81 & 13.07 & 15.05 \\ \midrule
        \multirow{8}{*}{BikeDC} & HAN & 25.40 ± 2.26 & 33.07 ± 5.16 & 40.00 ± 3.52 & 50.00 ± 3.94 & 49.58 ± 8.66 & 31.41 ± 9.42 \\ 
        ~ & MAGNN & 24.04 ± 3.53 & 32.10 ± 6.47 & 37.05 ± 4.20 & 49.41 ± 6.55 & 50.74 ± 7.35 & 29.74 ± 8.61 \\ 
        ~ & TensorGCN & 25.30 ± 2.68 & 33.99 ± 8.56 & 40.29 ± 3.73 & 50.58 ± 9.18 & 51.22 ± 7.08 & 31.47 ± 8.07 \\ 
        ~ & RTGNN  & 35.22 ± 7.25 & 48.25 ± 5.10 & 45.88 ± 3.76 & 57.05 ± 4.59 & 52.13 ± 7.97 & 31.40 ± 8.55 \\ 
        ~ & PTGB & 26.68 ± 9.45 & 35.73 ± 9.22 & 27.31 ± 9.32 & 36.35 ± 9.61 & 50.64 ± 7.54 & 31.48 ± 7.17 \\
        ~ & AdaSNN & 30.49 ± 7.14 & 40.73 ± 7.27 & 31.80 ± 7.33 & 41.42 ± 7.56 &  53.48 ± 6.68 & 31.02 ± 6.84 \\ 
        ~ & RSEF-MVGNN & \textbf{48.23 ± 6.80} & \textbf{62.16 ± 8.74} & \textbf{51.47 ± 5.46} & \textbf{63.82 ± 8.58} & \textbf{55.57 ± 6.85} & \textbf{32.35 ± 6.72} \\ 
        ~ & Gain & 13.04 & 13.91 & 5.59 & 6.77 & 2.09 & 0.87 \\ \midrule
        \multirow{8}{*}{PROTEINS} & HAN & 71.98 ± 0.59 & 76.04 ± 0.51 & 72.58 ± 0.62 & 76.83 ± 0.50 & 12.86 ± 2.48 & 14.43 ± 3.64 \\ 
        ~ & MAGNN & 72.19 ± 0.59 & 75.97 ± 1.25 & 72.66 ± 0.65 & 76.58 ± 1.24 & 13.12 ± 2.97 & 15.78 ± 2.86 \\ 
        ~ & TensorGCN & 72.19 ± 0.60 & 75.93 ± 0.69 & 72.75 ± 0.58 & 76.50 ± 0.67 & 12.98 ± 3.43 & 17.34 ± 3.12 \\ 
        ~ & RTGNN  & 73.28 ± 0.48 & 77.08 ± 1.18 & 73.81 ± 0.54 & 77.79 ± 1.19 & 14.77 ± 2.71 & 21.39 ± 2.56 \\ 
        ~ & PTGB & 65.37 ± 5.23 & 68.32 ± 4.67 & 66.14 ± 4.81 & 69.82 ± 4.65  & 15.84 ± 4.30 & 17.28 ± 4.27 \\
        ~ & AdaSNN & 73.68 ± 1.28 & 77.16 ± 1.03 & 73.22 ± 1.75 & 78.53 ± 1.86 & 19.53 ± 2.28 & 22.62 ± 3.48 \\  
        ~ & RSEF-MVGNN & \textbf{73.94 ± 1.40} & \textbf{78.34 ± 1.89} & \textbf{73.93 ± 1.22} & \textbf{79.39 ± 1.67} & \textbf{21.07 ± 2.19} & \textbf{24.47 ± 2.87} \\ 
        ~ & Gain & 0.26 & 1.18 & 0.12 & 0.86 & 1.54 & 1.85 \\  
        \midrule
        \multirow{8}{*}{ogbg-molhiv} & HAN & 63.39 ± 2.66 & 65.28 ± 2.55 & 65.52 ± 2.87 & 66.92 ± 2.74 & 14.20 ± 5.77 & 14.74 ± 5.62 \\ 
        ~ & MAGNN & 64.84 ± 2.88 & 66.34 ± 2.80 & 66.52 ± 3.12 & 67.28 ± 2.96 & 15.03 ± 3.65 & 15.46 ± 3.23 \\ 
        ~ & TensorGCN & 65.43 ± 2.33 & 66.64 ± 2.55 & 68.15 ± 2.84 & 70.09 ± 2.98 & 14.81 ± 3.83 & 15.72 ± 4.11 \\ 
        ~ & RTGNN  & 67.09 ± 4.61 & 68.17 ± 4.77 & 71.85 ± 5.10 & 72.33 ± 5.21 & 15.88 ± 3.44 & 17.58 ± 3.92 \\ 
        ~ & PTGB & 55.95 ± 3.34 & 58.40 ± 3.22 & 57.35 ± 3.15 & 59.03 ± 3.44 & 16.05 ± 4.84 & 16.49 ± 4.68 \\
        ~ & AdaSNN & 72.84 ± 4.45 & 74.93 ± 4.34 & 73.19 ± 4.28 & 75.55 ± 4.31 & 19.82 ± 3.95 & 21.47 ± 4.08 \\ 
        ~ & RSEF-MVGNN & \textbf{73.09 ± 3.14} & \textbf{75.53 ± 2.82} & \textbf{73.75 ± 3.11} & \textbf{76.08 ± 2.97} & \textbf{23.69 ± 4.55} & \textbf{26.77 ± 5.03} \\ 
        ~ & Gain & 0.25 & 0.60 & 0.56 & 0.53 & 3.87 & 5.30 \\ \midrule
        \multirow{6}{*}{ACM} & HAN & 91.06 ± 0.18 & 91.18 ± 0.24 & 91.11 ± 0.23 & 91.26 ± 0.36 & 61.56 ± 0.87 & 64.39 ± 0.95 \\ 
        ~ & MAGNN & 91.93 ± 0.25 & 92.17 ± 0.22 & 92.03 ± 0.22 & 92.34 ± 0.21 & 62.32 ± 1.06 & 64.77 ± 1.24 \\ 
        ~ & TensorGCN & 92.30 ± 0.33 & 92.49 ± 0.39 & 92.37 ± 0.35 & 92.52 ± 0.38 & 62.18 ± 1.13 & 64.05 ± 1.37 \\ 
        ~ & RTGNN  & 92.68 ± 0.37 & 92.88 ± 0.41 & 92.75 ± 0.39 & 93.07 ± 0.46 & 62.79 ± 0.92 & 65.18 ± 0.83 \\  
        ~ & RSEF-MVGNN & \textbf{93.19 ± 0.33} & \textbf{93.07 ± 0.35} & \textbf{93.28 ± 0.44} & \textbf{93.37 ± 0.42} & \textbf{64.15 ± 0.73} & \textbf{66.97 ± 0.84} \\ 
        ~ & Gain & 0.51 & 0.19 & 0.53 & 0.30 & 1.36 & 1.79 \\ 
        \bottomrule
    \end{tabular}
\end{table*}

\subsection{Experimental Settings}
\subsubsection{Dataset Details.}
1) \textbf{Human Immunodeficiency Virus (HIV)} \cite{HIV-} contains two types of imaging: functional magnetic resonance imaging (fMRI) and diffusion tensor imaging (DTI). Each instance includes DTI-derived brain graphs that share the same nodes with those derived from fMRI. 
2) \textbf{Capital Bikeshare Data (BikeDC)} \cite{transportation-} is collected from the Washington D.C. Bicycle System. Each instance is characterized by three temporal views: weekday, weekend, and monthly traffic patterns, which capture the complex traffic dynamics across 267 aggregated stations. 
3) \textbf{PROTEINS} \cite{protein-, ProData-} includes 1,000 protein molecule instances. Each instance contains two views: the sequence view and the molecule interaction view. 
4) \textbf{ogbg-molhiv} \cite{ogbg-} derived from MoleculeNet. Each instance in these datasets represents a molecule as a graph, where atoms are nodes and chemical bonds are edges. To address the scarcity of multi-view graph data for graph-level prediction tasks, we duplicate the original view to create a second view.
5) \textbf{ACM} \cite{HAN-} is derived from the ACM database, comprising two views: a co-paper graph (author-paper relationships) and a co-subject graph (subject-paper connections). 

\subsubsection{Compared Methods.}
1) \textbf{HAN} \cite{HAN-} aggregates relation-specific features from various graphs using attention techniques. 2) \textbf{MAGNN} \cite{MAGNN-} learns node attributes through linear transformation and aggregates information at a specialized encoder. 3) \textbf{TensorGCN} \cite{TensorGCN-} is a tensor GCN-based representation learning method for multi-view graphs. 4) \textbf{RTGNN} \cite{T2-} introduces a novel tensor GNN framework that enhances multi-view graph representation through reinforcement learning. 5) \textbf{PTGB} \cite{ptgb-} proposes a GNN pretraining framework for brain networks that captures intrinsic brain network structures. 6) \textbf{AdaSNN} \cite{R2-} proposes an adaptive subgraph neural network to detect critical structures in graphs.

\subsubsection{Implementation Details.}
Following the previous works \cite{T2-,MAGNN-}, we process the low-dimensional feature vectors generated by each method through Support Vector Machine (SVM) for classification tasks and through the K-means algorithm for clustering tasks. Specifically, for the SVM, we evaluate its performance across distinct training ratios, specifically at 20\% and 60\%, where the test set is provided to the linear SVM. In terms of clustering, the predefined number of classes in each multi-view dataset determines the number of clusters used in the K-means algorithm. 

\begin{figure}[t]
  \centering
    \includegraphics[width=1\linewidth]{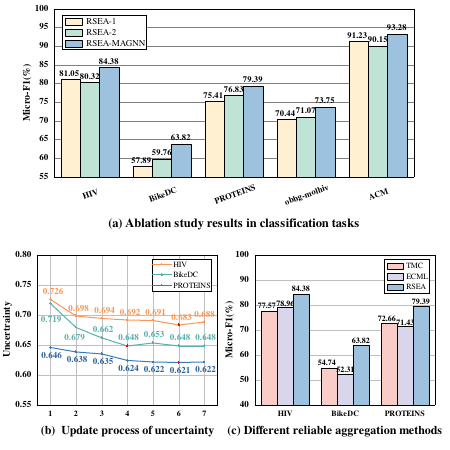}
  \caption{Ablation Study for RSEA-MVGNN.}
  \label{exp1}
\end{figure}

\subsection{Overall Performance (RQ1)}
Table \ref{expdata1} presents the performance comparison between RSEA-MVGNN and various baseline methods, with the best results highlighted in bold. Based on the results shown in Table \ref{expdata1}, we draw the following conclusions: 1) RSEA-MVGNN achieves top-performing performance across all five datasets,  improving classification by 13.91\% (BikeDC) and clustering by 15.05\% (HIV). 
2) RSEA-MVGNN notably improves weaker results in both classification and clustering tasks. On the HIV dataset, while PTGB performs well in classification (82.57\% Mi-F1), it shows poor results in clustering (34.67\% ARI). RSEA-MVGNN significantly enhances the weaker clustering results, improving both NMI and ARI by over 13\%. Similar improvements are seen in classification on BikeDC; clustering on PROTEINS, ogob-molhiv, and ACM datasets.
3) RSEA-MVGNN outperforms on five diverse datasets, while PTGB, a brain network-specific model, shows lower F1 scores on other domains. This highlights RSEA-MVGNN's effective generalization without domain-specific expert knowledge.

\subsection{Ablation Study (RQ2)}
Fig. \ref{exp1} (a) shows two RSEA-MAGNN variants: RSEA-1 lacks reliable structural enhancement, only enhancing a fixed number of priority nodes. RSEA-2 excludes the reliable aggregation in the inter-graph fusion. Neither variants of RSEA-MAGNN achieves the best performance. 
Fig. \ref{exp1} (b) visualizes the uncertainty reduction process, showing varying degrees of decrease across datasets during structural enhancement.
Fig. \ref{exp1} (c) shows classification tests comparing RSEA-MAGNN with reliable methods TMC \cite{SL2-} and ECML \cite{reliable-}. The results demonstrate that RSEA-MAGNN has advantage, indicating the necessity of introducing reliable mechanisms into GNNs for processing graph-type data.

\subsection{Computational Complexity Analysis (RQ3)}

\begin{figure}[t]
  \centering
    \includegraphics[width=\linewidth]{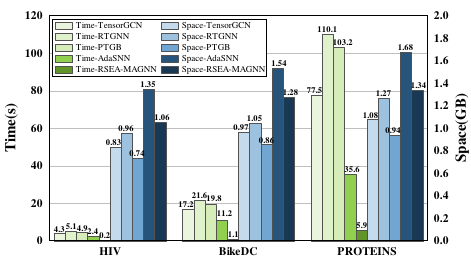}
  \caption{Execution Time (seconds) and Space Requirements (gigabytes).}
  \label{exp3}
\end{figure}

Fig. \ref{exp3} compares execution time and memory usage of five methods on various datasets, tested on an NVIDIA RTX 4090 GPU. RSEA-MVGNN shows significantly shorter execution time with comparable memory requirements.
The training process of RSEA-MVGNN including reliable structural enhancement with $\mathcal{O}(VMN^2)$, intra-graph aggregation with $\mathcal{O}(VN^2D)$, and inter-graph reliable aggregation with $\mathcal{O}(VNC^2)$. The overall time complexity can be represented as $\mathcal{O}(LVMN^2 + LVN^2D + LVNC^2)$, which simplifies to $\mathcal{O}(LVN^2(M + D))$. 

\section{Conclusion}
In this paper, we propose a novel multi-view GNN-based framework for representation learning, termed as RSEA-MVGNN. To enhance the graph's structural robustness and feature diversity, we design the reliable structural enhancement by feature de-correlation algorithm. For the purpose of enabling reliable aggregation in multi-view GNNs, we construct aggregation parameters, enabling high-quality views to dominate the inter-graph aggregation process. These two modules can be easily adapted to various GNN architectures. Experimental results show RSEA-MVGNN outperforms state-of-the-art baselines in multi-view graph neural networks.

\bibliography{aaai25}

\end{document}